\def\year{2020}\relax
\documentclass[letterpaper]{article} 
\usepackage{aaai20}  
\usepackage{times}  
\usepackage{helvet} 
\usepackage{courier}  
\usepackage[hyphens]{url}  
\usepackage{graphicx} 
\usepackage{amsmath}
\usepackage{amssymb}
\usepackage{amsthm}
\usepackage{booktabs}
\usepackage{multirow}

\usepackage{graphicx}  
\DeclareMathOperator{\softmax}{softmax}
\newcommand{\citet}[1]{\citeauthor{#1} \shortcite{#1}}
\newcommand{\citep}{\cite}

\title{Semantics- and Syntax-related Subvectors in the Skip-gram Embeddings (Student Abstract)}
\author{Maxat Tezekbayev, Zhenisbek Assylbekov, Rustem Takhanov\\ 
Department of Mathematics, School of Sciences and Humanities,
\\
Nazarbayev University, Nur-Sultan, Kazakhstan
}
 
\begin{document}

\maketitle

\begin{abstract}
We show that the skip-gram embedding of any word can be decomposed into two subvectors which roughly correspond to semantic and syntactic roles of the word.
\end{abstract}

\section{Introduction}
Assuming that words have already been converted into indices, let $\{1,\ldots,n\}$ be a finite vocabulary of words. Following the setups of the widely used {\sc word2vec} \cite{mikolov2013distributed} model, we consider \textit{two} vectors per each word $i$: 
\begin{itemize}
    \item $\mathbf{w}_i$ is an embedding of the word $i$ when $i$ is a center word,
    \item $\mathbf{c}_i$ is an embedding of the word $i$ when $i$ is a context word.
\end{itemize}  
We follow the assumptions of \citeauthor{assylbekov2019context} \shortcite{assylbekov2019context} on the nature of word vectors, context vectors, and text generation, i.e.
\begin{enumerate}
    \item \label{assump_1} A priori word vectors $\mathbf{w}_1,\ldots,\mathbf{w}_n\in\mathbb{R}^{d}$ are i.i.d. draws from isotropic multivariate Gaussian distribution:
    $\mathbf{w}_i\,\,{\stackrel{\text{iid}}{\sim}}\,\,\mathcal{N}\left(\mathbf{0},\,\textstyle{\frac1d}\mathbf{I}\right)$, where $\mathbf{I}$ is the $d\times d$ identity matrix. 
    \item Context vectors $\mathbf{c}_1,\ldots,\mathbf{c}_n$ are related to word vectors according to
    $\mathbf{c}_i=\mathbf{Qw}_i$, $i=1,\ldots,n$, for some orthogonal matrix $\mathbf{Q}\in\mathbb{R}^{d\times d}$. 
    \item \label{assump_3} Given a word $j$, the probability of any word $i$ being in its context is given by
\begin{equation}
p(i\mid j) \propto p_i\cdot{e^{\mathbf{w}_j^\top\mathbf{c}_i}}\label{eq:model}
\end{equation}
where $p_i=p(i)$ is the unigram probability for the word $i$.
\end{enumerate}

\begin{figure}
    \centering
    \includegraphics[width=.47\textwidth]{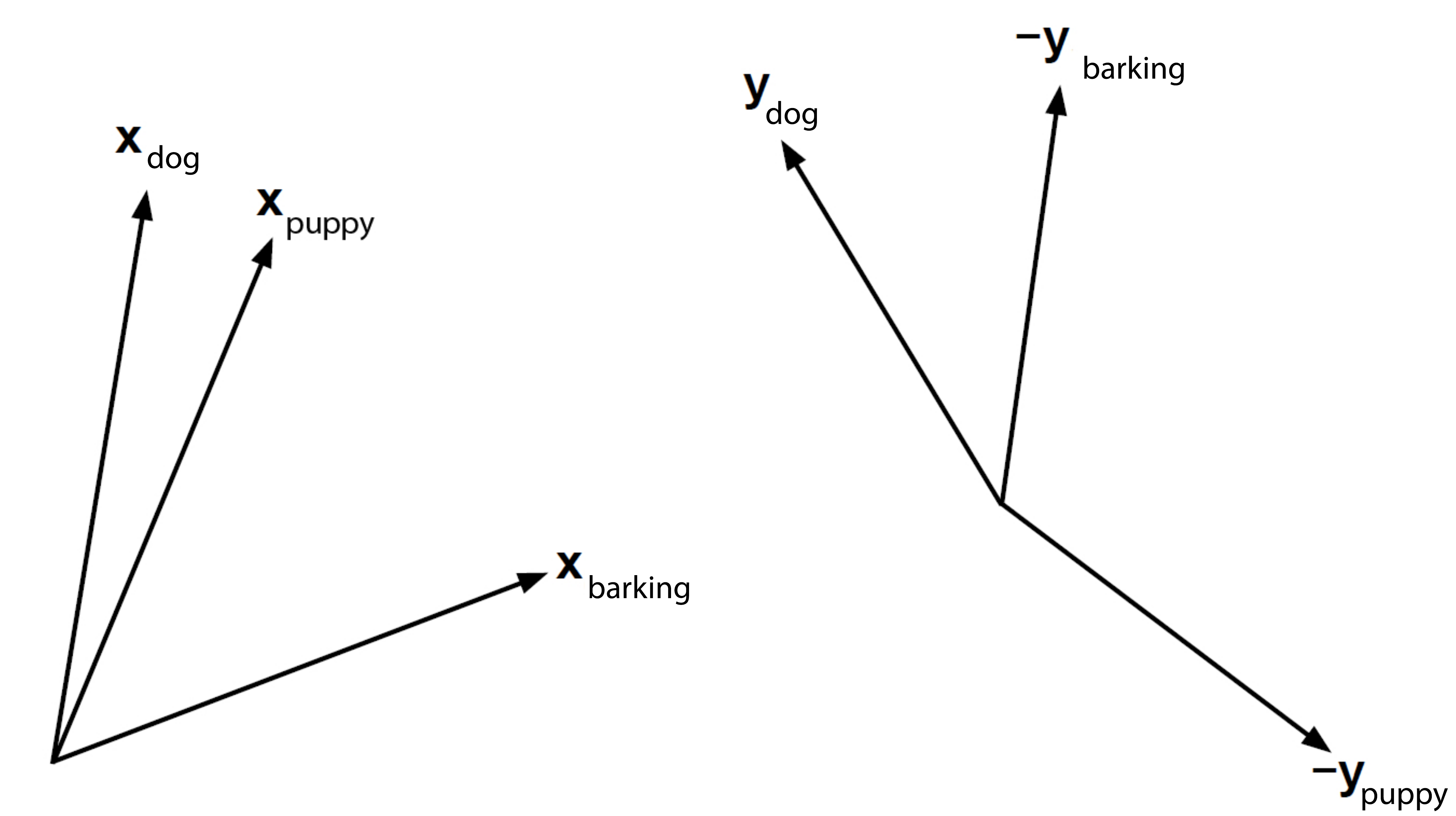}
    \caption{$\mathbf{x}$- and $\mathbf{y}$-embeddings}
    \label{fig:xy_embs}
\end{figure}

\noindent\textbf{Hypothesis}. Under the assumptions \ref{assump_1}--\ref{assump_3} above, \citeauthor{assylbekov2019context} \shortcite{assylbekov2019context} showed that each word's vector $\mathbf{w}_i$ splits into two approximately equally-sized subvectors $\mathbf{x}_i$ and $\mathbf{y}_i$, and the model \eqref{eq:model} for generating a word $i$ in the context of a word $j$ can be rewritten as
$$
p(i\mid j)\approx p_i\cdot e^{\mathbf{x}_j^\top\mathbf{x}_i-\mathbf{y}_j^\top\mathbf{y}_i}.
$$
Interestingly, embeddings of the first type ($\mathbf{x}_i$ and $\mathbf{x}_j$) are responsible for pulling the word $i$ into the context of the word $j$, while embeddings of the second type ($\mathbf{y}_i$ and $\mathbf{y}_j$) are responsible for pushing the word $i$ away from the context of the word $j$. We hypothesize that the $\mathbf{x}$-embeddings are more related to semantics, whereas the $\mathbf{y}$-embeddings are more related to syntax. In what follows we provide a motivating example for this hypothesis and then empirically validate it through controlled experiments.

\begin{table*}[htbp]
\centering
\begin{footnotesize}
\begin{tabular}{l | l c | c c c c | c c}
\toprule
\multirow{2}{*}{Data} & \multirow{2}{*}{Embeddings} & \multirow{2}{*}{Size} & Finkelstein et al.  & Bruni et al.  & Radinsky et al.  & Luong, Socher, and Manning & \multirow{2}{*}{Google} & \multirow{2}{*}{MSR} \\
 & & & WordSim & MEN  & M. Turk & Rare Words &      & \\
 \midrule
\multirow{3}{*}{\tt text8} & $\mathbf{w}:=[\mathbf{x};\mathbf{y}]$ & 200 & .646    & .650 & .636    & .063       & .305 & .319 \\
& Only $\mathbf{x}$ & 100 & .703 & .693 & .673 & .149 & .348 & .213 \\
& Only $\mathbf{y}$ & 100 & .310    & .102 & .193    & .019       & .032 & .128 \\
\midrule
\multirow{3}{*}{\tt enwik9} & $\mathbf{w}:=[\mathbf{x};\mathbf{y}]$ & 200 & .664 & .697 & .616 & .216 & .518 & .423 \\
 & Only $\mathbf{x}$ & 100 & .714 & .729 & .652 & .256 & .545 &.303  \\
 & Only $\mathbf{y}$ & 100 & .320 & .188 & .196 & .091 & .096 &.251 \\
 \bottomrule
\end{tabular}
\end{footnotesize}

\caption{Evaluation of word vectors and subvectors on the analogy tasks (Google and MSR) and on the similarity tasks (the rest). For word similarities evaluation metric is the Spearman's correlation with the human ratings, while for word analogies it is the percentage of correct answers. Model sizes are in number of trainable parameters.}
\label{tab:emb_eval}
\end{table*}

\section{Motivating Example}

Consider a phrase
$$
\textit{the dog barking at strangers}
$$
The word `barking' appears in the context of the word `dog' but the word vector $\mathbf{w}_{\text{barking}}$ is not the closest to the word vector $\mathbf{w}_{\text{dog}}$ (see Table~\ref{tab:proximities}). Instead, these vectors are split
\begin{align*}
    \mathbf{w}_{\text{dog}}^\top&=[\mathbf{x}_{\text{dog}}^\top; \mathbf{y}_{\text{dog}}^\top]\\
    \mathbf{w}_{\text{barking}}^\top&=[\mathbf{x}_{\text{barking}}^\top; \mathbf{y}_{\text{barking}}^\top]
\end{align*}
in such way that the quantity $\mathbf{x}_\text{dog}^\top\mathbf{x}_\text{barking}-\mathbf{y}_\text{dog}^\top\mathbf{y}_\text{barking}$ is large enough. We can interpret this as follows: the word `barking' is semantically close enough to the word `dog' but is not the closest one: e.g. $\mathbf{w}_\text{puppy}$ is much closer to $\mathbf{w}_\text{dog}$ than $\mathbf{w}_\text{barking}$; on the other hand the word `barking' syntactically fits better being next to the word `dog' than `puppy', i.e. $-\mathbf{y}_\text{dog}^\top\mathbf{y}_\text{puppy}<-\mathbf{y}_\text{dog}^\top\mathbf{y}_\text{barking}$.
\begin{table}[htbp]
    \centering
    \begin{footnotesize}
    \begin{tabular}{{c c c c c}}
    
    \toprule
    {word $i$} & {$\mathbf{w}_{\text{dog}}^\top\mathbf{c}_i$} & {$\mathbf{w}_{\text{dog}}^\top\mathbf{w}_i$} & {$\mathbf{x}_{\text{dog}}^\top \mathbf{x}_i$} &     {$-\mathbf{y}_{\text{dog}}^\top \mathbf{y}_i$} \\
    \midrule
    puppy & $-0.204$ & $13.331$ & $6.564$ & $-6.768$\\ 
    barking & $-0.263$ & $10.343$ & $5.040$	 & $-5.303$\\ 
    \bottomrule
    \end{tabular}
    \end{footnotesize}
    \caption{Dot products between vectors.}
    \label{tab:proximities}
\end{table}
This combination of semantic proximity ($\mathbf{x}^\top_\text{dog}\mathbf{x}_\text{barking}$) and syntactic fit ($-\mathbf{y}_\text{dog}^\top\mathbf{y}_\text{barking}$) allows the word `barking' to appear in the context of the word `dog'.

\section{Experiments}
In this section we empirically verify our hypothesis. We train SGNS with tied weights \cite{assylbekov2019context} on two widely-used datasets, \texttt{text8} and \texttt{enwik9},\footnote{\url{http://mattmahoney.net/dc/textdata.html}. The \texttt{enwik9} data was processed with the Perl-script {\sc wikifil.pl} provided on the same webpage.} which gives us word embeddings as well as their partitions:
$$\mathbf{w}_i^\top:=[\mathbf{x}_i^\top;\mathbf{y}_i^\top].$$
The source code that reproduces our experiments is available at \url{https://github.com/MaxatTezekbayev/Semantics--and-Syntax-related-Subvectors-in-the-Skip-gram-Embeddings}.

\subsection{$\mathbf{x}$-Subvectors Are Related to Semantics}
We evaluate the whole vectors $\mathbf{w}_i$'s, as well as the subvectors $\mathbf{x}_i$'s and $\mathbf{y}_i$'s on  standard semantic tasks --- word similarity and word analogy. We used the {\sc hyperwords} tool of \citeauthor{levy2015improving} \shortcite{levy2015improving} and we refer the reader to their paper for the methodology of evaluation. The results of evaluation are provided in Table~\ref{tab:emb_eval}. As one can see, the $\mathbf{x}$-subvectors outperform the whole $\mathbf{w}$-vectors in the similarity tasks and show competitive performance in the analogy tasks. However, the $\mathbf{y}$-parts demonstrate poor performance in these tasks. This shows that the $\mathbf{x}$-subvectors carry more semantic information than the $\mathbf{y}$-subvectors.

\subsection{$\mathbf{y}$-Subvectors Are Related to Syntax}
We train a softmax regression by feeding in the embedding of a current word to predict the part-of-speech (POS) tag of the next word:
$$
\widehat{\mathrm{POS}}[t+1] = \softmax(\mathbf{Aw}[t]+\mathbf{b})
$$
We evaluate the whole vectors and the subvectors on tagging the Brown corpus with the Universal POS tags. The resulting accuracies are provided in Table~\ref{tab:pos_eval}. 
\begin{table}[htbp]
    \centering
    \begin{footnotesize}
    \begin{tabular}{l c c c}
    \toprule
    \multirow{2}{*}{Embeddings} & \multirow{2}{*}{Size} & Trained on & Trained on \\
    & & \texttt{text8} & \texttt{enwik9} \\
    \midrule
    $\mathbf{w}:=[\mathbf{x};\mathbf{y}]$ & 200 & .445 & .453 \\ 
    Only $\mathbf{x}$ & 100 & .381 & .384\\ 
    Only $\mathbf{y}$ & 100 & .426 & .451\\ 
    \bottomrule
    \end{tabular}
    \end{footnotesize}
    \caption{Accuracies on a simplified POS-tagging task.}
    \label{tab:pos_eval}
\end{table}
We can see that the $\mathbf{y}$-subvectors are more suitable for POS-tagging than the $\mathbf{x}$-subvectors, which means than the $\mathbf{y}$-parts carry more syntactic information than the $\mathbf{x}$-parts.

\section{Conclusion}
Theoretical analysis of word embeddings gives us better understanding of their properties.  Moreover, theory may provide us interesting hypotheses on the nature and structure of word embeddings, and such hypotheses can be verified empirically as is done in this paper.

\section*{Acknowledgements}
This work is supported by the Nazarbayev University Collaborative Research Program 091019CRP2109, and by the Committee of Science of the Ministry of Education and Science of the Republic of Kazakhstan, IRN AP05133700.

\bibliography{ref}

\end{document}